\documentclass[fleqn,10pt]{wlscirep}
\usepackage[utf8]{inputenc}%
\usepackage[T1]{fontenc}%
\usepackage{bm}%
\usepackage{svg}%
\usepackage{graphicx}%
\usepackage{amsmath}%
\usepackage{subfiles} 
\usepackage[normalem]{ulem}
\usepackage[version=4]{mhchem}
\usepackage{hyperref}
\usepackage{siunitx}
\usepackage{lineno}

\let\oldflalign\flalign
\let\oldendflalign\endflalign

\renewenvironment{flalign}
{\linenomathNonumbers\oldflalign}
{\oldendflalign\endlinenomath}

\newif\ifsubmit
\submittrue
\definecolor{comments}{rgb}{0.1, 0.66, 0.1}
\ifsubmit
    \newcommand{\can}[1]{}
    \newcommand{\bo}[1]{}
    \newcommand{\tocite}[1]{}
    \newcommand{\todo}[1]{}
\else
    \newcommand{\can}[1]{[{\color{blue}CL: #1}]}
    \newcommand{\bo}[1]{[{\color{brown}Bo: #1}]}
    \newcommand{\tocite}[1]{[{\color{red}citation-#1}]}
    \newcommand{\todo}[1]{[{\color{red}TODO: #1}]}
\fi

\makeatletter
\renewcommand{\fnum@figure}{Fig. \thefigure \space|}
\makeatother

\title{Trustworthy Tree-based Machine Learning by \ce{MoS2} Flash-based Analog CAM with Inherent Soft Boundaries}

\author[1,$^{\dagger}$]{Bo Wen}
\author[1,$^{\dagger}$]{Guoyun Gao}
\author[1,2]{Zhicheng Xu}
\author[1]{Ruibin Mao}
\author[1]{Xiaojuan Qi}
\author[3]{X. Sharon Hu}
\author[2]{Xunzhao Yin}
\author[1,4,*]{Can Li}
\affil[1]{Department of Electrical and Electronic Engineering, The University of Hong Kong, Hong Kong SAR, China}
\affil[2]{College of Information Science and Electronic Engineering, Zhejiang University, Hangzhou, China}
\affil[3]{Department of Computer Science and Engineering, University of Notre Dame, Notre Dame, IN, USA}
\affil[4]{Center for Advanced Semiconductor and Integrated Circuit, The University of Hong Kong, Hong Kong SAR, China}
\affil[*]{E-mail: canl@hku.hk}
\affil[$^{\dagger}$]{These authors contributed equally to this work.}
\begin{abstract}
The rapid advancement of artificial intelligence has raised concerns regarding its trustworthiness, especially in terms of interpretability and robustness. 
Tree-based models like Random Forest and XGBoost excel in interpretability and accuracy for tabular data, but scaling them remains computationally expensive due to poor data locality and high data dependence. 
Previous efforts to accelerate these models with analog content addressable memory (CAM) have struggled, due to 
the fact that the difficult-to-implement sharp decision boundaries are highly susceptible to device variations, which leads to poor hardware performance and vulnerability to adversarial attacks.
This work presents a novel hardware-software co-design approach using \ce{MoS2} Flash-based analog CAM with inherent soft boundaries, enabling efficient inference with soft tree-based models. 
Our soft tree model inference experiments on \ce{MoS2} analog CAM arrays show this method achieves exceptional robustness against device variation and adversarial attacks while achieving state-of-the-art accuracy. 
Specifically, our fabricated analog CAM arrays achieve 96\% accuracy on Wisconsin Diagnostic Breast Cancer (WDBC) database, while maintaining decision explainability.
Our experimentally calibrated model validated only a 0.6\% accuracy drop on the MNIST dataset under 10\% device threshold variation, compared to a 45.3\% drop for traditional decision trees.
This work paves the way for specialized hardware that enhances AI's trustworthiness and efficiency.

\end{abstract}

\begin{document}

\flushbottom
\maketitle

\section*{Introduction}

In light of the rapid advancement of artificial intelligence (AI), concerns are escalating over its trustworthiness, specifically regarding minimizing the risk of unintended consequences by increasing decision explainability and the resilience against adversarial attacks. \cite{Floridi2019trustAI,arrieta2020explainable,liang2022advances,messeri2024artificial}.
Although deep neural networks (DNNs) based methods - including large language models (LLMs) - predominate modern AI approaches, their “black-box” nature and susceptibility to adversarial perturbations (minor, human-imperceptible input changes that drastically alter outputs) leading to questions about their trustworthiness.
\cite{gunning2019darpa,von2021transparency,quinn2022three}.
This limitation heavily restricts their use in high-risk areas such as healthcare \cite{loftus2020artificial}, legal systems \cite{vale2022explainable}, and unmanned transportation \cite{chen2024end}. 
On the other hand, tree-based models, such as Decision Tree (DT), Random Forest (RF) and XGBoost, offer inherent interpretability through human-readable decision rules while delivering accuracy that is comparable, or even surpasses that of deep learning models, especially for tabular data \cite{Grinsztajn2022tree_outperform,tree_tabular1,tree_tabular2}. 
However, scaling tree-based models remains challenging due to computational inefficiencies because of irregular memory access patterns (poor data locality) that lead to under-utilization of memory hierarchy and sequential decision dependencies (high data dependence) that complicate parallelizing operations on conventional digital hardware \cite{Aversa2024Leveraging}. Consequently, accelerating these models remains a complex task.

Emerging non-volatile memory (NVM) technologies, such as memristors or resistive memories \cite{ielmini2018memory,yao2020fullyCNN,rao2023thousands}, have opened up new possibilities for novel in-memory analog computing approaches beyond traditional von Neumann architectures by processing information at the exact location where data is stored, and massively parallel analog computation.
Most previous works for in-memory computing have focused on crossbar structures \cite{wan2022compute,ambrogio2023analog,yi2023activity}, which have demonstrated significant benefits in accelerating matrix multiplication in DNN, but offer limited value for tree-based models. 
Recent efforts to accelerate tree-based models leverage content-addressable memory (CAM), which is hardware that compares input data with stored data in rows parallelly and outputs the address of the match data. CAM represents one of the earliest commercial successes for in-memory computing. However, the SRAM-based CAM, which comprises 16 transistors in a cell, does not scale well for large-scale implementation. As a result, its use has been confined to specialized applications like cache memory and network routers, where high throughput is demanded, rather than capacity.
Recently, analog CAM has been proven to be effective for accelerating tree-based model inference \cite{Pedretti2021tree_aCAM_memristor,Yin2020FeCAM,Yin2024deepRF_FeCAM,pedretti2024x}, which fully exploited analog storage and multilevel capability of the memory device. Analog CAM stores data within the programmable conductance or threshold voltages as tunable continuous ranges and can take analog input as search values. 
Our previous 6T2R designs for analog CAM based on memristors \cite{Li2020aCAM,Pedretti2021tree_aCAM_memristor}
have demonstrated orders of magnitude improvement over conventional digital approaches in terms of TCAM (ternary CAM), but the soft searching boundaries and complex designs have limited the experimental demonstrations on physical arrays.
The more compact designs of 2-FET analog CAM based on ferroelectric field effect transistors (FeFET) \cite{Yin2020FeCAM,Yin2024deepRF_FeCAM} and floating gate transistors \cite{Danial2019_2D_floating_gate,Kazemi2021flash_MCAM,wu2021atomically,huang2023ultrafast} show even higher memory density and energy efficiency. Nevertheless, they exhibit an even softer boundary due to the physical limitations imposed by the sub-threshold swing. 
Therefore, existing implementations on analog CAM face two critical limitations. First, analog CAM cells inherently exhibit gradual conductance transitions rather than ideal step-function thresholds. This "soft" matching boundary prevents the implementation of the sharp decision boundary required by traditional hard tree models. Second, tree-based models are quite vulnerable to minor perturbations near decision boundaries. As a result, tree models implemented in analog CAM can be easily affected by threshold variation caused by device non-idealities such as threshold/conductance drifting and readout fluctuation, which emerging devices often suffer from \cite{Adam2018challenges_memristive,sun2021pcm,yon2022exploiting}. 
Most previous works have focused on sharpening the soft boundaries to store more levels in a memory cell, aiming to improve the precision \cite{Li2020aCAM,Pedretti2021tree_aCAM_memristor,Yin2020FeCAM,Yin2024deepRF_FeCAM}. But these optimizations made the tree models implemented in analog CAM even more sensitive with sharper boundaries, resulting in poor tolerance to threshold variation in hardware. This issue also increases the vulnerability of the model to adversarial attacks \cite{wei2024physical,chakraborty2021survey}, as small changes that are imperceptible to humans in input patterns can dramatically affect the final decision.

To address these challenges, we repropose the inherent \textit{soft boundaries} of analog CAM — viewed as a hardware limitation previously — to naturally implement soft tree models, such as soft decision trees (SDTs).
SDTs \cite{Irsoy2012SDT,Frosst_Hinton2017SDT} replaces the binary splits in traditional decision trees (DTs) with hard boundaries with probabilistic, sigmoid-like boundaries, showing improved performance while maintaining good interpretability.
However, soft tree models have not seen widespread adoption mainly due to their extremely high computation demands on conventional digital hardware. 
Specifically, they require evaluating all potential decision paths, and each decision node performs a complex nonlinear operation (e.g., sigmoid), leading to prohibitive inference complexity ($O(2^d)$ for depth-$d$ trees) rather than $O(d)$ for traditional decision trees.
In this work, we exploit the inherent soft boundaries of analog CAM, so that each analog CAM node naturally performs a sigmoid-like operation with physical laws, while the cells on the same row collectively compute the path probabilities in parallel. 
Our successful implementation is a result of the co-design of the hardware with natural soft boundaries and the algorithm with soft tree models.
Our hardware represents a different in-memory computing paradigm, where the memory cells are not only for multiplication but also for probabilistic computation.
The new hardware reduces the SDT inference latency from $O(2^d)$ to $O(1)$. 
Our simulation shows that CAM-implemented SDT is accelerated by $10^3\times$/$10^4\times$ compared to GPU/CPU-based solutions, with energy consumption improved by five to six orders of magnitude.  
On the software side, the probabilistic tolerance inherent to SDTs compensates for device non-idealities in analog CAM (such as conductance drift). 
This compensation effectively translates algorithmic robustness into hardware noise resilience, as validated by simulations based on an experimentally calibrated model: 
SDT accuracy on the MNIST dataset decreases by only 0.6\% compared to the ideal scenario, whereas traditional decision tree models suffer an average drop of 45.3\% under a 10\% device threshold variation.
As an extension, Soft Random Forest (SRF) also shows higher accuracy and robustness over the hard counterpart RF on the MNIST and tabular datasets similar to SDT versus DT.  
Furthermore, SDT models show better robustness against root-node adversarial attacks. SDT holds only a 1.7\% decrease in accuracy while DT suffers a 14.4\% drop.

We further demonstrate the SDT implementation on our fabricated 8×8 \ce{MoS2} charge-trapping Flash memory array. 
The two-dimensional (2D) \ce{MoS2} material is viewed as a promising material candidate to replace silicon in advanced nodes, because they offer several performance advantages, including high ON current, an excellent ON/OFF ratio, and low drain capacitance that leads to superior performance in analog CAM operation. 
Recent studies have also shown that 2D Flash can achieve fast and high-precision tuning of their memory states with low programming overhead \cite{bayat2016NORflash_analog,Danial2019_2D_floating_gate,Jiang2024ultrafast_2D_flash,Yu2023_2D_flash}. 
While our work focus on 2D Flash, the method is technology-agnostic and can be applied to other threshold tunable FET memories (e.g., FeFETs, Si-Flash, etc.).
Our SDT inference experiment on our physical \ce{MoS2} array achieves an accuracy of 96\% on the Breast Cancer dataset and 97\% on the Iris dataset, largely due to the significantly improved tolerance to device threshold variation, provided by the co-designed soft tree models and analog CAM hardware.
We further introduce a scalable analog CAM architecture to implement large soft tree models. 
Scalability is validated via circuit simulations of 1k-cell rows, showing <1\% mean error under real-world device variations.  
This work pioneers a paradigm shift: instead of forcing analog CAM to mimic digital precision, we exploit its native analog traits to enable high-performance, trustworthy AI. 
Our co-design approach bridges algorithmic robustness and hardware resilience, offering a pathway toward scalable and trustworthy machine learning.

\section*{Results}\label{sec2}
\subsection*{Soft tree models in analog CAM with soft boundaries}\label{sec2.1}












\begin{figure}[htp!]
    \centering
    \includegraphics[width=0.95\linewidth]
    {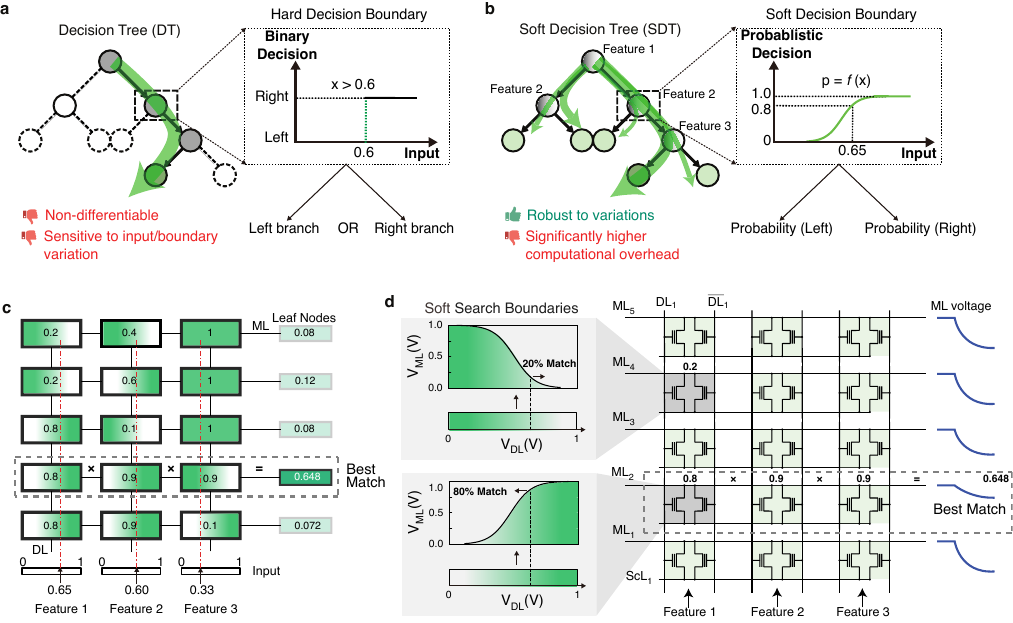}
    \caption{\textbf{Soft tree models in analog CAM with soft boundaries.} 
    \textbf{a}, Naïve Decision Tree (DT) models make each decision through a single path. The decision boundary in the node of the DT is sharp and is sensitive to perturbations around it.
    \textbf{b}, Soft tree models make decision based on the probabilities through all paths. The decision boundaries in inner nodes are soft and smooth which are more robust against perturbations, but the computation overhead is significantly increased.
    For a given input, multiple tree paths can be traversed with different probabilities, where the thickest green path indicates the path with the highest probability.
    \textbf{c}, Abstract analog CAM array implementing the SDT model in (b), where each root-to-leaf path corresponds to a row of the array. 
    Each node computes a nonlinear sigmoid-like probability rather than performing comparisons as in conventional CAM. 
    Each row aggregates the probabilities from the nodes along its path and outputs the highest probability as the decision.
    \textbf{d}, The circuit schematic for an analog CAM array with a compact 2-FET structure, where the match range is determined by the two threshold tunable FETs \cite{strachan2022analog,Yin2020FeCAM,Kazemi2021flash_MCAM}. 
    The searching boundaries of an analog CAM cell are gradual or soft, and the output on the ML when the cell is disconnected is essentially a nonlinear, sigmoid-like function of the input voltage, representing the probability of that node. 
    When the cells are connected, the final voltage on the ML results from the collective effect (such as the product) of the probabilities from all nodes in the row.
    The row that output the highest voltage on ML is the best match, indicating the leaf node with the highest probability in SDT.
    }\label{fig1}
\end{figure}

Unlike DT-based models with hard, axis-aligned decision boundaries (Fig. \ref{fig1}a), SDT \cite{Irsoy2012SDT} adopts a sigmoid-like decision boundary for each node, which denotes a partial probability for the final product of probabilities along the path (Fig. \ref{fig1}b). 
Fig. \ref{fig1}c illustrates our method for mapping a soft decision tree model onto an analog CAM.
In this method, each row of the analog CAM corresponds to a root-to-leaf path in the decision tree. 
Each cell within the CAM represents a node that calculates the nonlinear probability based on the input and the stored range. 
These node probabilities interact with each other, and the output of each row represents the overall probability for the corresponding root-to-leaf path.
The path that outputs the highest probability will be the decision of the tree. 
It is worth noting that this method differs fundamentally from previous mappings that used hard decision boundaries (Fig. S2 illustrates the comparison). 
In those cases, CAM performs a data lookup functionality, here each node performs a comparison operation that produces binary decisions based on definite sharp boundaries. 
This means that even minor variations in the device could lead to incorrect decision paths. The situation is exacerbated by the fact that the search boundaries in current analog CAM cell designs are not ideally sharp, further complicating the design of analog CAM hardware for accelerating tree-based models.

The circuit schematic of the analog CAM array is shown in Fig. \ref{fig1}d. 
In the design, each cell is composed of only two threshold-tunable FETs, which was previously invented by us 
\cite{strachan2022analog} and reported with ferroelectric FET\cite{Yin2020FeCAM} and floating-gate FET\cite{Danial2019_2D_floating_gate}.
The left and right FETs can adjust the high and low search boundaries, respectively, by tuning their threshold voltages. 
One drawback of these designs is the soft (non-abrupt) searching boundaries (inset in Fig. \ref{fig1}d), limiting the number of levels that can be stored in each cell. The inherent gradual switching nature in the transistors nature cannot precisely represent the ideal abrupt decision boundary in DT either. If efforts are paid to sharpen these soft boundaries, i.e., the switching characteristics of the transistors, another problem could be even more severe, i.e., minor distortions like device threshold variation in FET devices can trigger abrupt and significant changes around boundaries.
We observed that the output curve of the analog CAM cell with respect to the input is sigmoid-like, due to the softness in the searching boundary. 
Therefore, in this study, we leverage the inherent soft boundary of analog CAM cell, using it as a feature to execute the probability-based decision calculations in soft trees. 
Soft trees are much more tolerant to variations (Fig. \ref{fig1}b), as each inner node makes probabilistic decisions. 
Small changes in the input, or the decision boundaries change the continuous probability value, but many do not result in incorrect decisions. 

\begin{figure}[ht!]%
    \centering
    \includegraphics[width=0.95\linewidth]{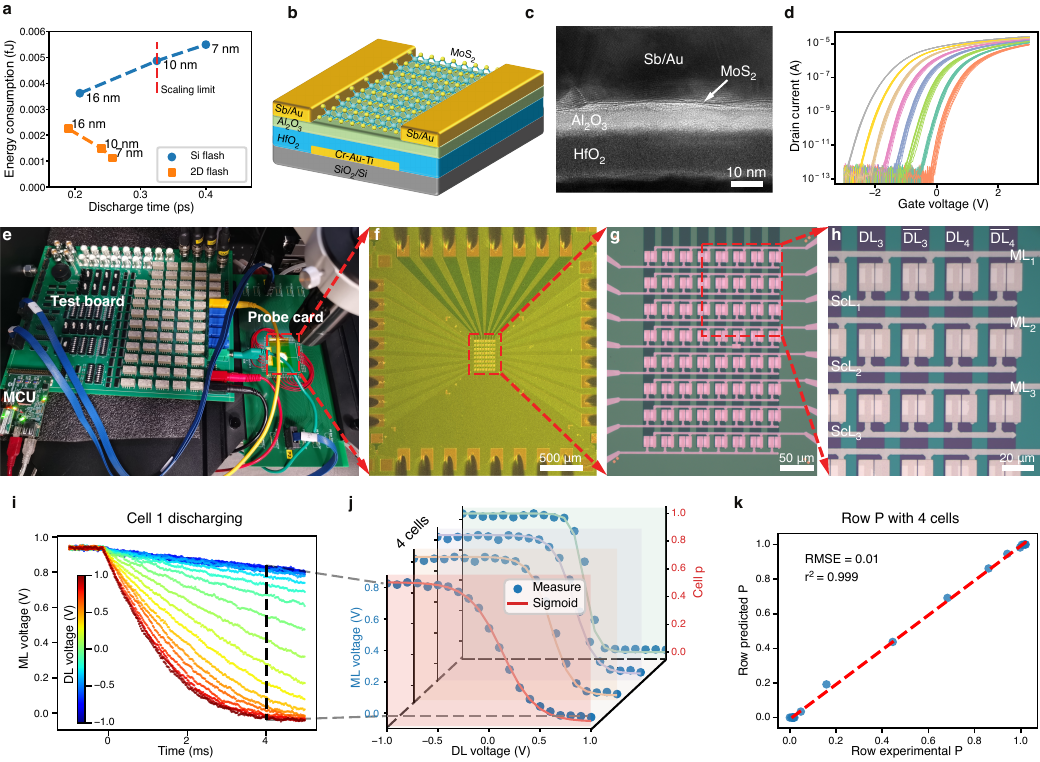}
    \caption{\textbf{Two-dimensional \ce{MoS2} Flash-based analog CAM arrays and operations} 
    \textbf{a}, The projected discharge time and energy consumption for flash memory utilizing Si-based and two-dimensional (2D) channel materials at advanced technology nodes \cite{Lu2024Projected_SI_2D}.
    \textbf{b}, Three-dimensional schematic of the 2D \ce{MoS2} Flash device. 
    \textbf{c}, Cross-sectional scanning transmission electron microscopy (STEM) image of our device, showing the \ce{Al2O3}/\ce{HfO2} as the charge-trapping gate dielectric, \ce{MoS2} channel material, and \ce{Sb/Au} for Ohmic contact. 
    \textbf{d}, The transfer characteristic curves of the device when programmed to eight different memory states, showing a high ON/OFF ratio of over $10^8$. 
    \textbf{e}, Our custom-built measurement platform comprises a PCB-based test board with a microcontroller (MCU) that communicates with a control computer. The test board is connected to a probe card in contact with 
    \textbf{f-h}, the 8 $\times$ 8 analog CAM array under test. 
    \textbf{i}, 
  Transient ML discharging curves for a single analog CAM cell, experimentally measured with DL input voltages swept from -1 V to 1 V.
    \textbf{j}, 
    ML voltage (sampled at 4 ms) versus DL input, fitted by modified sigmoid functions representing the cell’s probability $p$. Thresholds range stored in four cells of the same row are ``<-0.3V'',``<0.27V'',``<-0.75V'' and ``<0.77V''.
    \textbf{k}, 
    Experimental row output probability $P$ versus fitted predictions, resulted from a collective probability from each cell shown in (j), showing high agreement ($r^2>0.999, RMSE\approx 0.01$), corresponding to <1\% fitting error.
    }\label{fig2}
\end{figure}

\subsection*{Soft tree model inference experiment with 2D \ce{MoS2} analog CAM arrays}\label{sec2.2}

To validate the idea, we fabricated $8\times8$ analog CAM arrays using 2D \ce{MoS2} charge-trap Flash memory for soft tree model inference experiments. 
We chose 2D \ce{MoS2} Flash devices because the 2D material can provide higher ON/OFF ratio than silicon-based devices due to better gate controllability, and low drain capacitance due to the atomic thin thickness. 
This benefit is more apparent in advanced technology node, which is crucial for the speed and energy efficiency of the flash CAM discharges/searches \cite{Lu2024Projected_SI_2D}.
As shown in Fig. \ref{fig2}a, projected discharging time and energy consumption for 2D-based Flash significantly outperform silicon-based devices at scaled nodes, because silicon Flash faces fundamental scaling limit below 10 nm \cite{ishimaru2021challenges,wang2022road}. 
Moreover, silicon Flash-based analog CAM may suffer from increasing drain capacitance, since a thinner gate oxide separates the gate and the source/drain electrode aggravates the effects of intrinsic and parasitic capacitance \cite{kim2012high}. 
In addition to the benefit during the inference, the use of 2D Flash also helps lower the latency and energy to reconfigure the memory states and thus the model. 
It was reported that the programming speed increases dramatically from 10–100 $\mu$s in silicon Flash to 10–20 ns in 2D devices \cite{liu2021ultrafast, Yu2023_2D_flash, Jiang2024ultrafast_2D_flash}.
It is noted that the soft tree mapping idea is also applicable to other more mature threshold-tunable transistors, such as ferroelectric FETs and non-2D Flash memories, but could exhibit a slightly higher latency and energy due to material constraints.

The schematic of our \ce{MoS2} Flash memory device is illustrated in Fig. \ref{fig2}b and the cross-sectional STEM (scanning transmission electron microscopy) image is shown in Fig. \ref{fig2}c.
More details of device fabrication can be found in \hyperlink{methods}{Methods}.
The devices demonstrate a high ON/OFF ratio (over $10^8$)), and a wide tuneable memory window, as shown in the transfer characteristic curves in Fig. \ref{fig2}d. These curves show at least eight non-overlapping programmable states, and they can be tuned continuously in a range. 
To support the tree model inference experiment, we built a PCB-based system, as shown in Fig. \ref{fig2}e, to supply the voltages to program/read devices, circuit to pre-charge/discharge the match lines (MLs) for model inference, and sensing circuit to collect the ML voltage.
The system is controlled by a micro-controller unit (MCU) connected to the general-purpose computer.
More details about the system can be found in Fig. S3.
Fig. \ref{fig2}f, g\&h shows an $8\times8$ analog CAM array under test with probe card. 
The paired data lines (DLs) are connected through analog inverters on the test board during model inference. The selection of columns and rows is also realized on the test board using multiplexers and relays. 

We first validate that each analog CAM cell can perform a sigmoid-like function. 
In the experiment, the low boundary of a single cell is programmed to \SI{-0.3}{\volt} to implement “$<\SI{-0.3}{\volt}$”, while the other cells in the same row are programmed to ``X'' to match all inputs.
The transient discharging curve sensed from the ML with varying input voltages from -1V to 1V is showed in Fig. \ref{fig2}i. 
As expected, the ML stays high when the input voltage is much smaller than \SI{-0.3}{\volt}, while it quickly discharges if the input is much larger than \SI{-0.3}{\volt}.
If we extract the ML voltage sensed at a predefined time point during the search (e.g., \SI{4}{\milli\second} in Fig. \ref{fig2}j, or other time points in Fig. S4),
we find that the transition between match and mismatch is smooth and continuous. 
As mentioned above, this creates a challenge because it limits the number of levels that can be stored in each cell and increases complexity when mapping a hard tree model to analog CAM. 
We also observe that the ML voltage discharging curves of 4 cells in a row can be well fitted by sigmoid functions (Equation \ref{eq2}), as shown in Fig. \ref{fig2}j. And the softness of the curve can be adjusted by sense latency (as showed in Fig. S4). 
In the soft decision tree experiment we have conducted later, we use the sigmoid function result to represent the probabilities $p_i$ of the cell/node as shown below,
\begin{flalign}
    p_i &= \sigma(k(V_{\text{in},i}- V_{\text{th},i})) \label{eq2}
\end{flalign}
where $\sigma$ represents the sigmoid function and $k$ is a gain to represent the softness of the soft curves. 
$V_{\text{in},i}$ and $V_{\text{th},i}$ are the input voltage on DLs and the stored thresholds in analog CAM cells. 

The final decision of the soft tree model is made based on the root-to-leaf path with the highest probability, which is the product of the probabilities of each node along the path in the original algorithm.
The analog CAM array can also calculate a similar probability product based on the probability ($p_i$) using the sigmoid-like function of each cell.
In order to simulate the discharging behavior of the ML voltage in an analog CAM row, which represents the path probability product $P$, we introduce the behavior model below,
\begin{flalign}
    V_{\text{ML},t} = P =a \prod^n_i p_i + b \sum^n_i p_i - b(n-1)V_{{ML,t_0}}, 0 \le P \le 1 & \label{eq1} 
\end{flalign}
where $a$ and $b$ are fitting parameters. We set hard limits during the fitting to ensure that $P$ remains within the range of $[0,1]$. 
It is mainly the probability product of cells along the path, with two additional terms of summation and compensation based on the number of cells $n$ and the precharged ML voltage $V_{{ML,t_0}}$.
The detailed circuit analysis that leads to the equation can be found in 
\hyperlink{methods}{Methods}. 
It is noted that the ML voltage is not strictly a product of probabilities from each cell, but rather a combination of the product and the sum of probabilities.
We observe that for a large analog CAM array, the coefficient $b$ tends to be quite small (close to 0), and therefore the result is closer to the product of probabilities.
Even for a small array, we found out that as long as we train the SDT model based on the known relationship, the performance of SDT model is not affected by the sum term of probabilities in Equation \ref{eq1}.
Fig. \ref{fig3}k shows that the model matches the experimental results well  in physical hardware. 
The predicted values of $P$ obtained from our model are highly correlated with the values of the experiment on a row with 4 cells, with a high coefficient of determination ($r^2$) of 0.999. The low RMSE (root mean square error) of about 0.01 also indicates a fitting error low to 1\%. 

\subsection*{Breast cancer diagnostic experiment}

\begin{figure}[hpt!]
  \centering
  \includegraphics[width=0.95\linewidth]
  {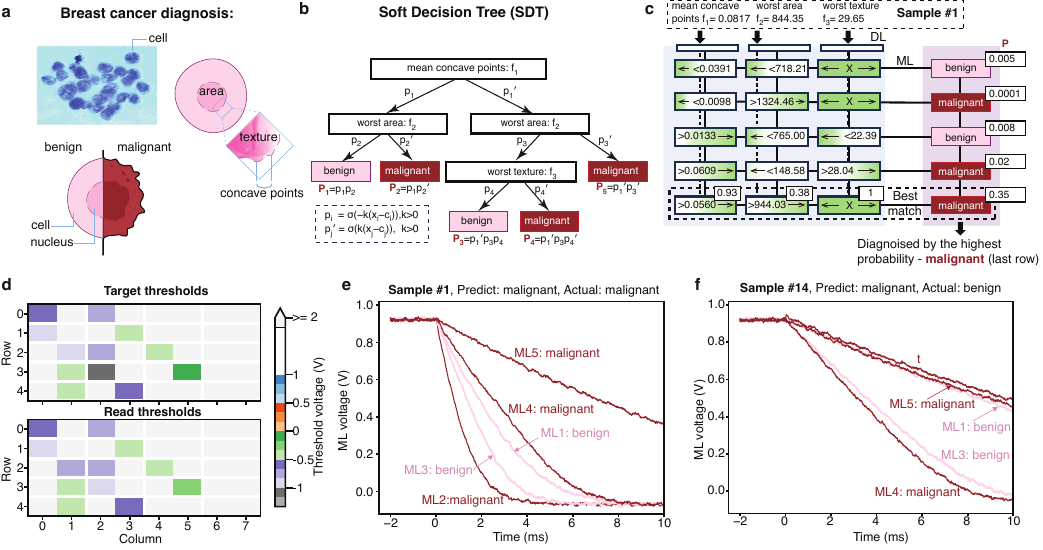}
  \caption{\textbf{Beast cancer diagnostic experiment with soft decision tree in analog CAM arrays.}
  \textbf{a}, A sample microscope image of a breast mass sample, and the cell features extracted from the images (e.g., texture, area, concave points) drawn from the Breast Cancer Wisconsin (Diagnostic) database \cite{breast_cancer_wisconsin}, for benign/malignant diagnosis.
  \textbf{b}, 
  Soft decision tree (SDT) sharing the Decision Tree (DT)'s structure but using probabilistic sigmoid nodes and leaves encoding path probabilities. Thresholds are initialized from the DT and fine-tuned via gradient descent; classification selects the highest-probability leaf.
  \textbf{c}, The trained SDT that mapped to the two-dimensional arrays of analog CAM.
  Each analog CAM cell calculates the sigmoid-like probability based on the input feature and the stored range, and the paths's probabilities are calculated by the product of the probabilities from each node along the path.
  The feature values are normalized to the operation range of the analog CAM, and the thresholds are programmed to the corresponding voltages in (d).
  A detail circuit schematic of the analog CAM is shown in Fig. S6.
\can{Remove the WTA label, and replace all numbers here to the raw numbers before matching. You can add a circuit schematic with the WTA circuit, etc., in the SI if you have one already.}
  \textbf{d}, 
The mapping target (upper) and the actual (lower) threshold voltages after programming the analog CAM devices. White indicates a threshold voltage high enough ($\geq 2V$) for the ``always match'' condition. The threshold voltage is defined as the gate voltage when the drain current reaches 10 nA.
\textbf{e, f},
Examples of experimentally observed transient ML discharging curve. 
The curves showing dark red color corresponds to the decision path for malignant diagnosis, where light red color indicates the benign diagnosis.
(e) shows the correct diagnosis with input features shown in (c), where the ML5 (path 5) discharges the slowest and thus has the highest probability.
(f) shows an example of incorrect diagnosis, where the ML1 and ML5 discharges at similar rates, indicating similar probabilities for both paths, so that a human expect can inspect this case. 
  }\label{fig3}
\end{figure}

After validating that our analog CAM array can effectively perform SDT inference, we deploy a software-hardware co-designed model on the hardware to demonstrate breast cancer diagnosis.
The diagnosis determines whether a breast cancer sample is benign (not cancer) or malignant (cancer) based on features extracted from its digital microscope images from the Breast Cancer Wisconsin (Diagnostic) database \cite{breast_cancer_wisconsin} (Fig. \ref{fig3}a).
A baseline DT is trained using standard \textit{scikit-learn} python package \cite{scikit-learn} (Fig. S5), and the three most important features are selected via cost complexity parameter optimization: \textit{mean concave points} (quantifying nucleus irregularity, as the average number of concave portions of nuclei contours), \textit{worst area} (a measure of the largest nucleus size, by pixel count), and \textit{worst texture} (characterizing chromatin distribution patterns, as the largest variance of gray-scale intensities).  
After that, the SDT is trained based on the behavior model described in Equation \ref{eq1} (from the previous section), and using the exactly the same features and tree structure as the DT.
More details about the SDT and DT training is shown in \hyperlink{methods}{Methods}.
The trained SDT model is illustrated in Fig. \ref{fig3}b.

The trained SDT model is then mapped to the analog CAM array, where each row corresponds to a root-to-leaf path in the tree.
The input features and hardware threshold voltage are normalized to the range of [-1, 1].
Previous implementations to map conventional, hard DT use identical threshold values for split nodes shared across multiple tree paths, exactly following the software models \cite{Pedretti2021tree_aCAM_memristor,Yin2020FeCAM}.
This is reasonable for the hardware tree, because when traversed through the nodes in the tree structure, only one path can be selected at a time, and the other paths are not used.
For soft tree, all paths are traversed simultaneously, so the mapping to analog CAM can allow the split nodes to have different threshold values for different paths.
This approach enables optimized mapping of SDT paths onto individual rows of the analog CAM array, where features are assigned column-wise (Fig. \ref{fig3}c).
Instead of rigidly replicating software-defined node thresholds, our method allows each CAM row to implement a path-specific component of a soft decision boundary \can{I remember we had a discussion and determined to use decision boundry, or soft decision boundry instead of threshold, to distinct from hardware threshold.} . 
This decoupling from a single, global threshold per split node leverages the analog CAM's full programmability to enhance model fidelity in hardware.
These soft decision boundaries are then implemented in the analog CAM by programming the threshold voltages of the two FETs in each cell, which are linearly mapped the range of [\SI{-1}{\volt}, \SI{1}{\volt}], since the thresholds of our 2D charge-trapping FETs can be reliably and precisely programmed in this range.
Fig. \ref{fig3}d shows the experimentally readout threshold values after programming the analog CAM array, compared to the target values.
The results show small programming error in threshold voltages in the analog CAM array, with a minor difference which is no greater than \SI{0.1}{V}.
The devices in unused cells are simply programmed to a typical high threshold (e.g. $\geq$\SI{2}{V}) for the "always match".
The mapping of "less than" and "larger than" is realized by programming the left and right devices in a cell with the corresponding threshold voltages.
More details are described in \hyperlink{methods}{Methods}.

After the SDT model is trained and deployed on our analog CAM array, the diagnosis is performed experimentally.
The MLs of the analog CAM array are first precharged to a high voltage to prepare for the inference.
The input features are then applied to the DLs of the array (where each feature is represented by a pair of DLs connected via an analog inverter), causing the MLs to discharge according to how well the input matches each path's conditions.
The ML voltage, representing the probability product of the corresponding path in the SDT model, is sampled and read out at a predefined time point after inference begins.
The row with the highest ML voltage represents the selected path for the diagnosis result, which can be hardware readout using a winner-takes-all (WTA) circuit attached to the MLs.
Experimentally, out of 143 samples in the dataset, 137 were correctly diagnosed as benign or malignant, achieving an accuracy of 95.8\% (Fig. \ref{fig3}e).
This accuracy surpasses that of the hard DT version implemented in digital hardware (93.7\% accuracy), correctly diagnosing three additional samples.\can{Please double check these numbers.}
Furthermore, the deployed SDT model retains good explainability, as it preserves the tree structure and path probabilities, thereby allowing for human interpretation.
Fig. \ref{fig3}c and Figs. \ref{fig3}e-f provide an example of how samples are diagnosed using the model and hardware.
The experimentally recorded discharging curves from the MLs (Figs. \ref{fig3}e, f) show that MLs discharge at different rates; curves that discharge slower will be sampled at a higher voltage, representing a higher probability for the corresponding path.
(The discharging rate was intentionally slowed by the capacitance of the off-chip measurement boards to facilitate observation of the discharging curves using an oscilloscope; with on-chip peripherals, this process could be several orders of magnitudes faster.)
Fig. \ref{fig3}e illustrates a case where ML5 (the leaf node for path 5) discharges the slowest. 
This indicates the highest probability for this path, leading to a correct diagnosis of malignant for a sample characterized by high values for mean concave points in its cells and a large worst area. 
Conversely, Fig. \ref{fig3}f shows an example of an incorrect diagnosis, where several discharging curves are close, indicating similar probabilities for multiple paths.
In such instances, human experts can inspect the contributing factors. 
For this incorrect prediction, the path leading to the erroneous result (malignant) is primarily influenced by a large worst area. 
The expert (e.g., a doctor) can also observe that an alternative path (path 4, in this example) has a similarly high probability. 
This alternative path, characterized by small mean concave points despite a large worst area, would suggest a benign diagnosis. 
In this situation, the doctor can integrate this nuanced output with other clinical information to make a more informed and accurate diagnosis.
To further validate its generality, we reconfigured the analog CAM for classification on the Iris dataset (Fig. S8), achieving 97\% classification accuracy with only one misclassified sample.


\subsection*{Variation and attack resilience of soft tree models in analog CAM}\label{sec2.3}

\begin{figure}[ht!]
  \centering
  \includegraphics[width=0.95\linewidth]
  {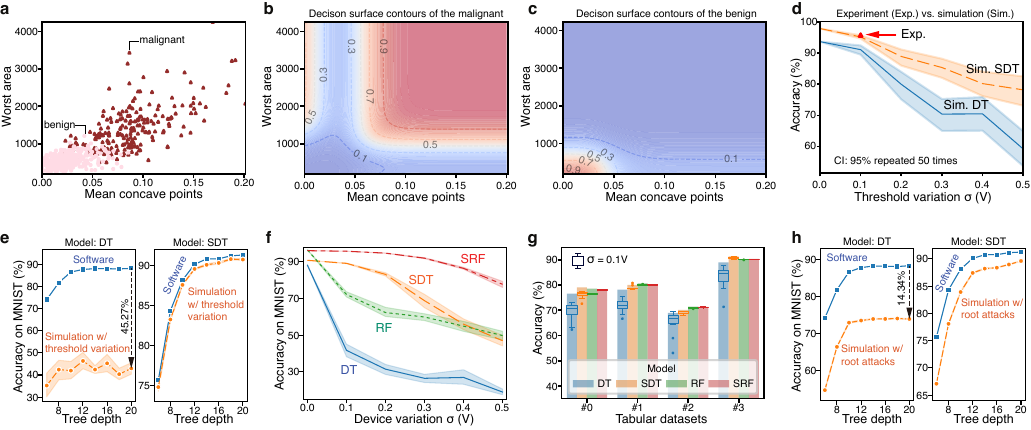}
  \caption{\textbf{Analysis of the robustness of soft tree models in analog CAM against device threshold variation and adversarial attack.} The DT/RF used for comparison shares the exact tree structure and feature regularization of SDT/SRF, as well as the same row/column architecture for implementation. The main difference is the mapping thresholds of soft models are different through the training considering the soft boundaries in an SDT way. 
  \textbf{a}, The malignant and benign samples for WDBC dataset with two features: \textit{mean concave points} and \textit{worst area}.
  \textbf{b}, The decision surface contours and heatmap of probability for the malignant. The feature not shown is fixed to be an averaged value.
  \textbf{c}, The decision surface contours heatmap of probability for the benign.
  \textbf{d}, Simulated accuracy (95\% CI) under device threshold variations (50 trials). SDT shows stronger robustness and achieves 95.8\% hardware accuracy in previous real device experiment (vs. 97.9\% software), outperforming the DT baseline (93.7\% software, no variation).
  \textbf{e}, The model's accuracy increase steadily with the rise of tree depth. SDT shows much better robustness under the device variation with only a loss of 0.6\% of accuracy when the tree depth is 20, while the significant drop for DT  is 45.2\%. 
  \textbf{f}, The simulation results of models' accuracy on MNIST with the same tree depth of 16 under normal distributed device threshold variation with increasing $\sigma$.
  \textbf{g}, The simulation results of model accuracy on four tabular datasets (\#0 electricity, \#1 bankmarketing, \#2 default-of-credit-card-clients and \#3 MiniBooNE). The boxes inside the bars are results under device threshold variation with a $\sigma$ of \SI{0.1}{\volt}. 
  \textbf{h}, SDT also shows stronger robustness than DT under the root node adversarial attack with only 1.7\% decrease compared to the 14.3\% drop for DT.
  }\label{fig4}
\end{figure}

\can{Why do you add this sentence here? seems irrelevant.- \sout{The DT implementation adopted for comparison is exactly the same way as SDT with the same tree structure.} }
\bo{That's something to answer editor's 2nd concern at first. }
We believe that one of the reasons behind the superior accuracy results for the CAM-based SDT implementation is the robustness of the soft tree model against device variation. 
Device variation is usually the main challenge for analog computing hardware, especially those based on emerging memory devices.
Fortunately, soft tree models implemented in analog CAM exhibit greater noise resilience compared to their hard tree counterparts, because soft tree models, by design, handle perturbations around the soft decision boundary more effectively, and our training process takes into consideration the gradual decision boundary observed in our experiments.
This significantly improved robustness of SDT against device variation is illustrated using the Breast Cancer diagnosis experiment in Fig. \ref{fig4}a-c.
Fig. \ref{fig4}a shows the distribution of benign and malignant samples based on the features \textit{mean concave points} and \textit{worst area}. 
Traditional hard DTs employ orthogonal decision boundaries, making diagnosis near these hard boundaries highly sensitive to hardware-induced shift due to device variation. 
In contrast, the SDT model utilizes soft decision boundaries, represented as complex surfaces where the probability changes gradually (illustration in Fig. S9).
The contours and probability heatmap of these decision surfaces are shown in Fig. \ref{fig4}b and c. 
The decisions do not change abruptly but rather shift smoothly in terms of probability as variation occurs.
Consequently, a small perturbation around the decision boundary will not significantly affect the final decision, leading to better robustness.
To evaluate robustness for different magnitudes of device variation, we simulated model accuracy across 50 Monte Carlo trials under increasing device threshold variations, presented with a 95\% confidence interval (CI) in Fig. \ref{fig4}d.  
These results clearly demonstrate SDT's superior resilience compared to DT as the variation magnitude increases, with the simulation well matches our experiments on real device.
Simulations on the Iris dataset (Fig. S10a, b) with both uniform and Gaussian variation distributions reveal similar trends: while both DT and SDT accuracy degrade with increasing variation, SDT consistently maintains significantly higher robustness. 
A comparable analysis of decision boundaries for the Iris dataset (Fig. S11) further explains this robustness across different datasets.

The performance and robustness of DT/SDT are further evaluated on the MNIST handwritten digit recognition dataset \cite{lecun2010mnist} and several tabular datasets \cite{Grinsztajn2022tree_outperform}. 
Before implementing these models in hardware, we compared their software accuracy to baseline Random Forest (RF) models (see Table.S2 in supplementary information). 
Our results show that the accuracies achieved by our implementations are on par with those reported in the literature \cite{Grinsztajn2022tree_outperform,Random_Forest} for similar model configurations.
While tree-based models are not typically strong performers in MNIST, it provides a valuable benchmark for quantitative comparison with other technologies and demonstrate the robustness of the proposed soft tree model.
All robustness results presented are averaged over 10 repetitions to account for randomness. 
The accuracy increase with trees' depth in the absence of device variation, as shown in Fig. \ref{fig4}e. 
As expected, SDT consistently outperforms DT, at a depth of 20, the accuracy of DT and SDT is 88.3\% and 91.3\%, respectively.
More importantly, the SDT model is much more robust against device threshold variation.
To evaluate this, we apply a uniformly distributed threshold variation (ranging from \SI{-0.1}{\volt} to \SI{0.1}{\volt}, about 10\% of the range, which is close to the experiment case). 
As presented in Fig. \ref{fig4}e, DT models experience a significant drop in accuracy, falling to less than 50\%, an accuracy drop of 45.27\% compared with the variation-free case.  In contrast, SDT models demonstrate superior resilience, with their accuracy almost unaffected, showing a mere decrease of 0.6\% at the same tree depth. 

The advantages of soft boundaries extend to ensemble methods like Random Forest (RF). 
Our soft boundaries method is orthogonal to the ensemble learning method, and can be easily extended to the soft version of RF (soft RF, or SRF) by replacing the hard boundaries with soft ones to achieve even better performance.
To evaluate this, we add the normal distributed threshold variation to the RF model and its soft version, SRF, and compare their performance with DT and SDT against device threshold variation.
All the models are with the same tree depth of 16 and soft tree models share the same tree structure with their hard versions. 
As illustrated in Fig. \ref{fig4}f, both hard and soft random forest models achieve over 96\% accuracy with 50 trees, outperforming single decision tree models when no variation is present. 
The soft versions of the models exhibit greater robustness against threshold variation compared to their hard counterparts; notably, SDT even surpasses hard RF models at slight device variation. 
When the device variation standard deviation reaches \SI{0.1}{\volt}, similar to our experimental conditions, the accuracy drop for SRF and SDT is only 0.3\% and 1.7\%, respectively, whereas the drop for RF and DT is significantly larger, at 24.3\% and 46.4\%.


Beyond the MNIST benchmark, we assessed robustness on four tabular datasets (``electricity'', ``bankmarketing'', ``default-of-credit-card-clients'', and ``MiniBooNE''), representing real-world applications where tree models excel \cite{Grinsztajn2022tree_outperform} and trustworthiness is paramount. 
In fields like finance, energy, and scientific research (represented by these datasets), model instability can lead to financial losses, infrastructure breakdowns, or flawed research outcomes. 
Therefore, robustness against perturbations like sensor noise or device variation is critical for reliable decision-making.
All the models we used for these datasets have a depth of six, and forest models consist of 250 trees. 
Fig. \ref{fig4}g shows the accuracy of the models on these datasets with and without device threshold variation.
The bars with black square markers represent the models' accuracy without variation, while the boxes within each bar depict their performance with device threshold variation (with standard deviation of \SI{0.1}{\volt}). 
The result further validated that soft tree models consistently outperform their hard counterparts with the presence of device variation.
For instance, in dataset \#3 "MiniBooNE" (to distinguish electron neutrinos from muon neutrinos), the variation effect is more pronounced, and SRF demonstrates greater robustness compared to RF and other models. 
In the other datasets, SRF and RF show similar performance, regardless of variation. 
Consistent with the MNIST dataset, SDT can sometimes outperform the forest models, as observed in dataset \#2 "default-of-credit-card-clients". 

Another benefit that soft tree models bring is their robustness against adversarial attacks.
To evaluate this, we subject both DT and SDT to an adversarial attack by applying a uniform distribution from 0 to \SI{1}{V} to their root nodes, which are typically considered the most vulnerable points in tree models.
This evaluation is performed on the MNIST dataset for a quantitative comparison.
Specifically, we replace only one pixel of the input image with random noise at the same order of magnitude. 
As shown in Fig. \ref{fig4}h, our findings reveal that DT models are highly vulnerable to the adversarial attack and experience a significant decrease in accuracy, with the maximum accuracy barely exceeding 75\%. 
In contrast, SDT models are much more stable due to their soft boundaries. Under root node adversarial attack, SDT models even outperform DT models that are not attacked when the depth is over 16. 
Additionally, the robustness of SDT models against adversarial attacks gradually increases with the tree depth. The SDT model with a depth of 20 performs the best under attack, with only 1.7\% decrease in accuracy, while the drop for DT is 14.3\%.

\subsection*{Scalable architecture and performance benchmarks}

\begin{figure}[hpt!]
  \centering
  \includegraphics[width=0.95\linewidth]{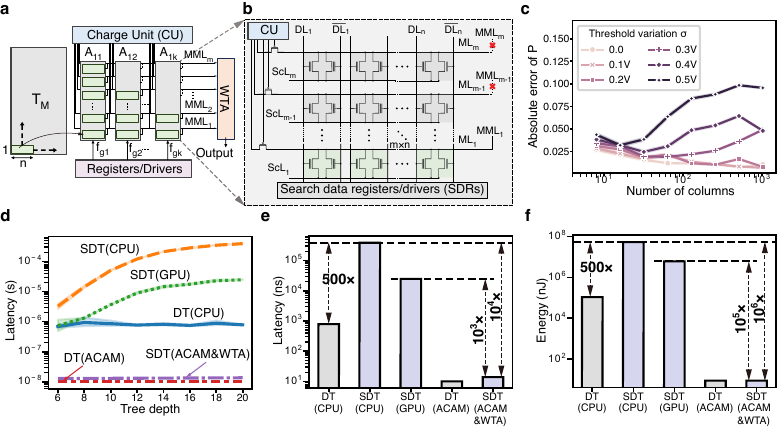}
  \caption{\textbf{Scalable architecture and performance benchmarks.}
  \textbf{a}, The scalable architecture of analog CAM  with $k$ sub-arrays to map the target values of a large model ($T_M$). Each inner ML inside sub-arrays (grey boxes) can independently connect or disconnect to the highway master MLs (MMLs). The charge unit is able to charge every sub-array separately.
  \textbf{b}, The architecture of sub-arrays in (a). 
  \textbf{c}, The error between circuit simulation and the proposed behavior model (Equation \ref{eq1}) for analog CAM array with increasing column number under device threshold variation with increasing standard deviation $\sigma$. 
  \textbf{d}, The inference latency of DT and SDT with the rise of tree depth for each sample implemented in analog CAM (ACAM) compared to GPU (RTX 3090) and CPU (AMD EPYC 7413). 
  \textbf{e} \& \textbf{f}, The inference latency and energy consumption of DT and SDT with a tree depth of 20 for each sample.
  }\label{fig5}
\end{figure}

To directly deploy soft tree models in real-world scenarios using an analog CAM array, a very large array would be required. 
However, this is impractical due to device variations and the sparsity during the mapping of tree models in analog CAM. 
To overcome these challenges, we propose a scalable analog CAM architecture for implementing soft tree models, as illustrated in Fig. \ref{fig5}a. 
The threshold array after mapping, represented as $T_M$ in Fig. \ref{fig5}a, is generally sparse. 
By reordering the features, we can place the most important ones—those frequently used in decision-making—on the left. 
The paths that require more features to reach the leaf nodes are also reordered and positioned lower in the array. 
This allows us to divide the large array into $k$ subarrays with a width of $n$. 
Only the subarrays located in the lower left part of the array are populated, therefore we can disable the unused subarrays and thereby improve the speed and energy.
These techniques are similar to the previously proposed hard tree model implementation in analog CAM \cite{Pedretti2021tree_aCAM_memristor}, but the soft tree model implementation presents a different challenge.
In the previous implementation, the results read from the MLs of analog are binary, indicating either a match (correct path), or a mismatch (incorrect path) in the tree/forest model.
This requirement limits the number of columns in each tile of the analog CAM, because it is necessary to distinguish between a match and a one-bit (or few-bits) mismatch. 
However, using smaller array tile sizes can solve this problem, and the tiles can be connected horizontally using simple digital logic of local decisions in the small tiles for global decisions.
In contrast, our design outputs analog values representing match degrees, which makes storing and processing outputs of small tiles (narrow arrays) more challenging and resource-intensive. 
Fortunately, our method does not distinguish between match and mismatch; instead it picks the highest match degree, a continuous value, and therefore, can support a much larger array compared to the binary case.
So, in our architecture, instead of connecting different tiles with digital logic, we connect the MLs in the subarrays to a Master ML (indicated as MML in Fig. \ref{fig5}a).
Only the enabled subarrays are connected to the MMLs, and the final result is obtained from the MMLs through a winner-take-all (WTA) circuit to determine the path with the highest match degree. 
Although we only show the subarrays organized in horizontally in Fig. \ref{fig5}a, the architecture can be easily scaled vertically with cascading WTA circuits for more decision paths.
To validate our assumption that our design can support a larger array, we conducted simulations to assess the accuracy of the analog CAM array as the number of columns increases. 
In the SPICE simulation, we extracted transistor models and wire resistance from our measurements of 2D material devices and arrays. 
The result is shown in Fig. \ref{fig5}c, depicting the absolute error of the output probability $P$, read from the MLs, as the absolute value of the mean error between the circuit simulation and the ideal probability calculation (Equation \ref{eq1}). 
We can see that the probability error first decreases with the number of columns (the number of cells in a row) 
, due to averaging of threshold variation. 
However, beyond a certain point, the probability error increases significantly due to various non-ideal factors.
Generally, with a large threshold variation 
, the analog CAM array can be scaled larger while maintaining a small error. 
When the standard deviation of the threshold variation is \SI{0.1}{\volt}, which is close to the value we observed in our experiments, the analog CAM array can sustain an error of less than 0.01 with up to 1,000 columns.

Finally, we evaluate the performance of the scalable analog CAM architecture in simulation for implementing a soft decision tree model, considering a complete circuit includes the analog CAM array and ML pre-charging circuit. 
In the analysis below, we use MNIST dataset 
as an example, and the inference batch size is set to 10,000, with values averaged over 10 repetitions. 
The tree models performance can be improved by increasing the tree depth, but that comes with a significant increase in computational complexity, leading to much longer inference latency and higher energy consumption in traditional digital hardware.
This is demonstrated by an exponential growth in latency with increasing tree depths for SDT on both CPU and GPU shown in Fig. \ref{fig5}d.
The slowdown of the latency growth observed beyond the tree depth of 12 is due to stronger effects of tree pruning, which reduce the model size.
Our benchmark results, shown in Fig. \ref{fig5}d
, illustrate a different trend for tree models implemented in analog CAM, both soft and hard models. 
The latency remains nearly constant with tree depth, and is orders of magnitude faster than the digital implementations.
This is because the sigmoid function and the multiplication of probabilities can be performed in the analog CAM discharging process efficiently as we mentioned above. 
Fig. \ref{fig5}e\&f compare the latency and energy consumption of DT and SDT models with the tree depth of 20. Both models share the same structure, featuring over 3,000 paths at this depth. The result shows that the computationally intensive SDT, when implemented in analog CAM, can achieve a remarkable speedup, with 3-4 orders of magnitude improvement compared to GPU (RTX 3090) and CPU (AMD EPYC 7413). The energy consumption per MNIST sample is significantly lower for SDT implemented in analog CAM, at 8.78 nJ and 0.07 nJ consumed in arrays and the WTA circuit\cite{WTA}. This represents a reduction by five and six orders of magnitude respectively when compared to GPU and CPU solution. The benchmark data with related works is shown in Table.S4.

\section*{Discussion}\label{sec5}
In summary, this study harnessed the inherent soft boundaries of analog CAM cells to enhance the accuracy, robustness, and interpretability of tree models executed within analog CAM. The co-designed analog CAM hardware facilitated computationally intensive sigmoid and probability calculations in parallel and within the analog domain, directly where the tree models are stored. Soft tree models, as opposed to basic hard tree models, exhibited improved interpretability, higher accuracy, and notably, exceptional robustness against device threshold variation. Even in the presence of about 10\% hardware threshold variation, the SDT maintained nearly full accuracy with only a marginal 0.6\% decrease, showcasing a stark contrast to the 45.3\% drop observed in the DT model on the MNIST dataset. Through the fabrication of 8$\times$8 arrays utilizing 2D materials as a proof of concept, the demonstration of a SDT model achieved an impressive accuracy of 96\% on the WDBC dataset and 97\% on the Iris dataset in the presence of device non-idealities. 
The collaborative efforts in emerging electronic device development, unconventional circuit and hardware design, and model development mark a pioneering step toward developing specialized hardware optimized for trustworthy and efficient machine learning and artificial intelligence applications.

\section*{Methods}
\label{methods}
\hypertarget{methods}{}
\subsection*{Device fabrication of \ce{MoS2} Flash memory arrays}
The fabrication of the 2D \ce{MoS2} flash memory device starts with deposition of back gate on substrate. 3 nm Cr, 10nm Au, and 1 nm Ti are used as the back gate. After oxygen plasma treatment, \ce{HfO2} (15 nm) and \ce{Al2O3} (5 nm) are then deposited with atomic layer deposition (ALD) as blocking layer and tunneling layer. Afterwards, mono-layer \ce{MoS2} growed by CVD is transferred by PMMA (Poly-methyl Methacrylate) and DI (Deionized) water, patterned by photo lithography and reactive ion etching (RIE). Finally, Sb (20 nm) and Au (30 nm) are deposited as contact electrodes for low contact resistance.

\subsection*{Circuit simulation}
2-FET analog CAM cell and arrays are simulated in LTspice and Ngspice. The device's simulation parameters including $KP$, $\lambda$ are obtained by fitting outputs of circuit simulation with experiment device transfer \& output curves. 
A custom python script generates the netlist using Pyspice for analog CAM arrays with different size (rows and columns number) and arbitrary device thresholds and input voltages. 

\subsection*{Circuit model for calculating probability with analog CAM}
A row of cells can carry out a probability or likelihood calculation. 
First, for FET devices connected to the same ML and shared the same drain voltage $V_d$, there are mainly 2 cases:
\begin{itemize}
    \item If $V_{g_i}<V_{{th}_i}$, the device works in the subthreshold region and the drain current $I_d$ is very small and negligible.
    \item If $V_{g_i}>V_{{th}_i}$, it may work in the saturation region at first and then fall into the linear region with decreased $V_d$ (or directly in the linear region)
\end{itemize}
If we let $V_{ov_i}=V_{g_i}-V_{{th}_i}>0$, the drain current $I_d$ can be expressed as:
\begin{flalign}
&& V_d \ge V_{ov_i}: &I_{d,i} = \frac{\text{KP}}{2}\frac{W}{L} V_{ov_i}^2(1+\lambda V_d) & \label{eqId1} \\
&& V_d < V_{ov_i}: &I_{d,i} = \text{KP}\frac{W}{L} (V_{ov_i}-\frac{V_d}{2})V_d(1+\lambda V_d) & \label{eqId2}
\end{flalign}
where $KP, \lambda, \frac{W}{L}$ are constant parameters of transistors.
Then the dynamic process of the pull-down transistors working in saturation region ($V_d \ge V_{ov_i}$) or linear region ($V_d < V_{ov_i}$) can be described by:
\begin{flalign}
&& V_d \ge V_{ov_i}: &\frac{dV_d}{dt} = -\frac{1}{C_{ML}} \sum_i \frac{1}{2}k V_{ov_i}^2(1+\lambda V_d) & \label{eq5} \\
&& V_d < V_{ov_i}: &\frac{dV_d}{dt} = -\frac{1}{C_{ML}} \sum_i k (V_{ov_i}-\frac{V_d}{2})V_d(1+\lambda V_d) & \label{eq6}
\end{flalign}
where $k$ is a constant parameters, and $C_{ML}$ is the ML capacitance. 
For Eq. (\ref{eq5}), we can easily get its solution as:
\begin{flalign}
    && V_d &= (V_{{ML,t_0}}+\frac{1}{\lambda})e^{-\frac{\lambda}{C_{ML}}\sum_i \frac{1}{2}kV_{ov_i}^2t}-\frac{1}{\lambda} & \label{eq7}
\end{flalign}
where $V_{{ML,t_0}}$ is the initial precharged voltage of ML. 
The solution is closely related to the exponential form that satisfy $e^{\sum_i V_{ov_i}^2}=\prod_i^ne^{V_{ov_i}^2}$.
The solution for Eq. (\ref{eq6}) could be very complicated. However, if we consider either a very small $\Delta t$ or negligible changes in $V_d$, an approximate solution can be derived, associated with another exponential form which satisfy $e^{\sum_i (V_{ov_i}-\frac{V_d}{2})}=\prod_i^ne^{(V_{ov_i}-\frac{V_d}{2})}$. 
This approximation enables us to utilize the $V_{ov_i}$ of each cell to predict the $V_d$ which is also the ML voltage in a product form. 
To achieve this, we assign the probability of a single cell as $p_i$, which can be calculated using the modified sigmoid function Equation \ref{eq2}.
Consequently, the ML voltage of a row can be expressed as the product of these probabilities, such as $P \approx \prod p^2_{i}$ or $P \approx \prod p_i$.

On the other hand, if the time interval $\Delta t$ is sufficiently long or ML voltage changes significantly, we can derive the following equations from a single cell to a row consisting of $n$ cells: 
\begin{flalign}
    &&  \text{Single cell: }&  p_i = V_{d}  = V_{d,t_0}-\frac{I_{d} \Delta t}{C_{ML}} & \label{eq8} \\
    &&  \text{Row with cells: }&  V_{ML}  = V_{{ML,t_0}}-\sum^n_i \frac{I_{d_i} \Delta t}{C_{ML}} = \sum^n_i p_i - (n-1)V_{{ML,t_0}} & \label{eq9}
\end{flalign}
Therefore, the ML voltage of the entire row can be obtained from the individual probabilities $p_i$ of each cell in a sum form, resulting in $P = \sum^n_i p_i - (n-1)V_{{{ML,t_0}}}$.
To summarize, the behavior of the ML voltage in an analog CAM row can be described as a product of individual cell probabilities ($p_i$) based on $V_{ov,i}$ at the beginning of the searching process when $V_\text{ML}$ drops slightly, and as a sum when the ML voltage drops near zero. 
To capture this behavior, we propose the simplified model Equation \ref{eq1} for easier co-optimization of the SDT algorithm.

\subsection*{Training algorithm of tree models}
All hard tree-based models are trained in a Python environment with Scikit-learn package. The soft tree models are trained using gradient descent techniques based on their hard counterparts with the same tree structure, in particular to feature selection for each node, every decision path of the tree, and the distribution in leaf nodes. The structure is already pruned during the training of hard tree models. 
We have discovered that the thresholds in inner nodes are repeatedly mapped in different rows of analog CAM. 
Therefore, we train the node thresholds shared by multiple decision paths by assigning them with the same initial values and then tuning them to change individually. This approach can further unlock the potential of analog CAM without incurring additional hardware overhead. Moreover, this approach is equivalent to transforming the binary SDT into a multi-way SDT with enhanced modeling capabilities.
For Iris dataset, we randomly split the training/testing set as 120/30. The maximum tree depth is 3 for DT/SDT. The inputs are regularized to [-1,1] first. 
For MNIST dataset, the pixel inputs are normalized from 0-255 to 0-1. All soft tree models are trained within a fixed epoch number. 

\subsection*{Differences between mapping ``smaller than'' and ``larger than'' soft thresholds}
\begin{itemize}
    \item For "$<$" case: The soft threshold of the left device in the analog CAM cell represents the upper bound. Therefore, We directly map the upper bound to the left device. The right one is programmed to a high threshold for "X". 
    \item For "$>$" case: Since the inputs for the two devices in a cell are connected with an analog inverter, we map the lower bound to the right device in the analog CAM cell with its opposite value. For example, "$>0.41 V$" needs to be mapped to the right device of the cell with the threshold voltage of "$-0.41 V$". The threshold of the left one in the cell is programmed high for "X".
\end{itemize}



\section*{Data availability}
The experiments of devices and arrays within this paper are measured through Keysight B1500 and the measurement platform we built. Associated source data for Figs. \ref{fig2},\ref{fig3},\ref{fig4},\ref{fig5} in this paper are provided with Figshare.

\section*{Code availability}
The code that supports the results in this paper is available on GitHub at \url{https://github.com/carlwen/CAM-SoftTree}.

\nolinenumbers
\bibliography{sn-bibliography}



\end{document}


\makeatletter
    \def\@maketitle{%
  \newpage
  \null
  \vskip 2em%
  \begin{center}%
  \let \footnote \thanks
    {\LARGE \@title \par}%
    \vskip 1.5em%
    {\large
      \lineskip .5em%
      \begin{tabular}[t]{c}%
        \@author
      \end{tabular}\par}%
  \end{center}%
  \par
  \vskip 1.5em}
\makeatother

\captionsetup{labelsep=space}
\makeatletter
\renewcommand{\fnum@figure}{Fig. \thefigure \space|}
\makeatother
\linespread{1.5}

\title{Supplementary Information 

Trustworthy Tree-based Machine Learning by \ce{MoS2} Flash-based Analog CAM with Inherent Soft Boundaries}

\author[1,$^{\dagger}$]{Bo Wen}
\author[1,$^{\dagger}$]{Guoyun Gao}
\author[1,2]{Zhicheng Xu}
\author[1]{Ruibin Mao}
\author[1]{Xiaojuan Qi}
\author[3]{X. Sharon Hu}
\author[2]{Xunzhao Yin}
\author[1,4,*]{Can Li}
\affil[1]{Department of Electrical and Electronic Engineering, The University of Hong Kong, Hong Kong SAR, China}
\affil[2]{College of Information Science and Electronic Engineering, Zhejiang University, Hangzhou, China}
\affil[3]{Department of Computer Science and Engineering, University of Notre Dame, Notre Dame, IN, USA}
\affil[4]{Center for Advanced Semiconductor and Integrated Circuit, The University of Hong Kong, Hong Kong SAR, China}
\affil[*]{E-mail: canl@hku.hk}
\affil[$^{\dagger}$]{These authors contributed equally to this work.}

\maketitle
\flushbottom
\newpage

\begin{figure}[ht!]%
    \centering
    \includegraphics[width=9cm]{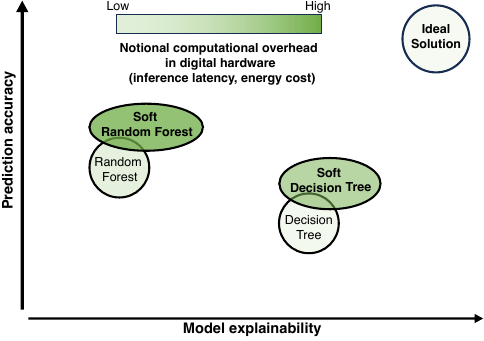}
    \caption{\textbf{The prediction accuracy, model explainability, and computational overhead of several tree-based models.}
    The shade of color represents the notional computational overhead in digital hardware. The soft tree-based models providing higher accuracy and explainability with probability-based decisions are inevitably accompanied by much heavier computational overhead than common tree-based models. 
}\label{tree_complexity}
\end{figure}

\begin{figure}[ht!]%
    \centering
    \includegraphics[width=12.5cm]{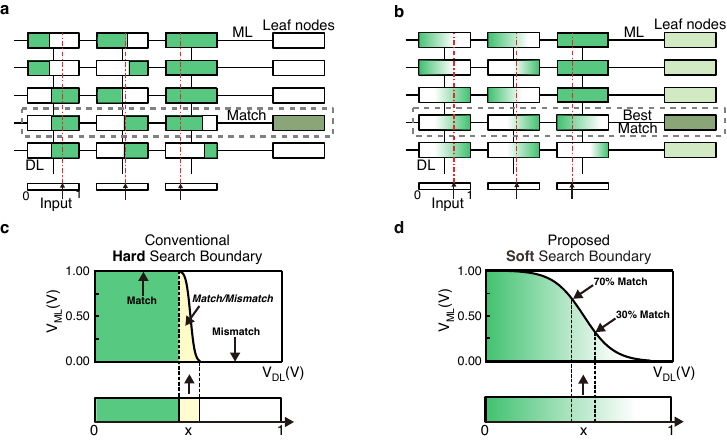}
    \caption{\textbf{Comparison of proposed soft tree mapping with previous hard tree mapping.} 
    \textbf{a}, Analog CAM array performing the traditional hard DT, where each root-to-leaf path corresponds to a row of the array. The selected path is represented by the parallel searching result reflected on ML, as the only one match result.
    Tree-based models can be mapped and accelerated in analog CAM arrays. The data lines (DLs) accept input in form of voltages and output the results through match lines (MLs).
    \textbf{b}, Analog CAM array performing SDT, where the boundary are soft. The output is the best match leaf with the highest probability.
    \textbf{c}, The conventional search boundary in analog CAM is expected to be sharp enough for binary match or mismatch outputs. The boundary is not ideally orthogonal with inevitable intermediate state.
    \textbf{d}, The proposed search with soft boundary outputs continuous results with degrees of match rather than binary match or mismatch.
    }\label{DT_SDT_comparision}
\end{figure}

\clearpage

\begin{figure}[ht!]%
    \centering
    \includegraphics[width=9cm]{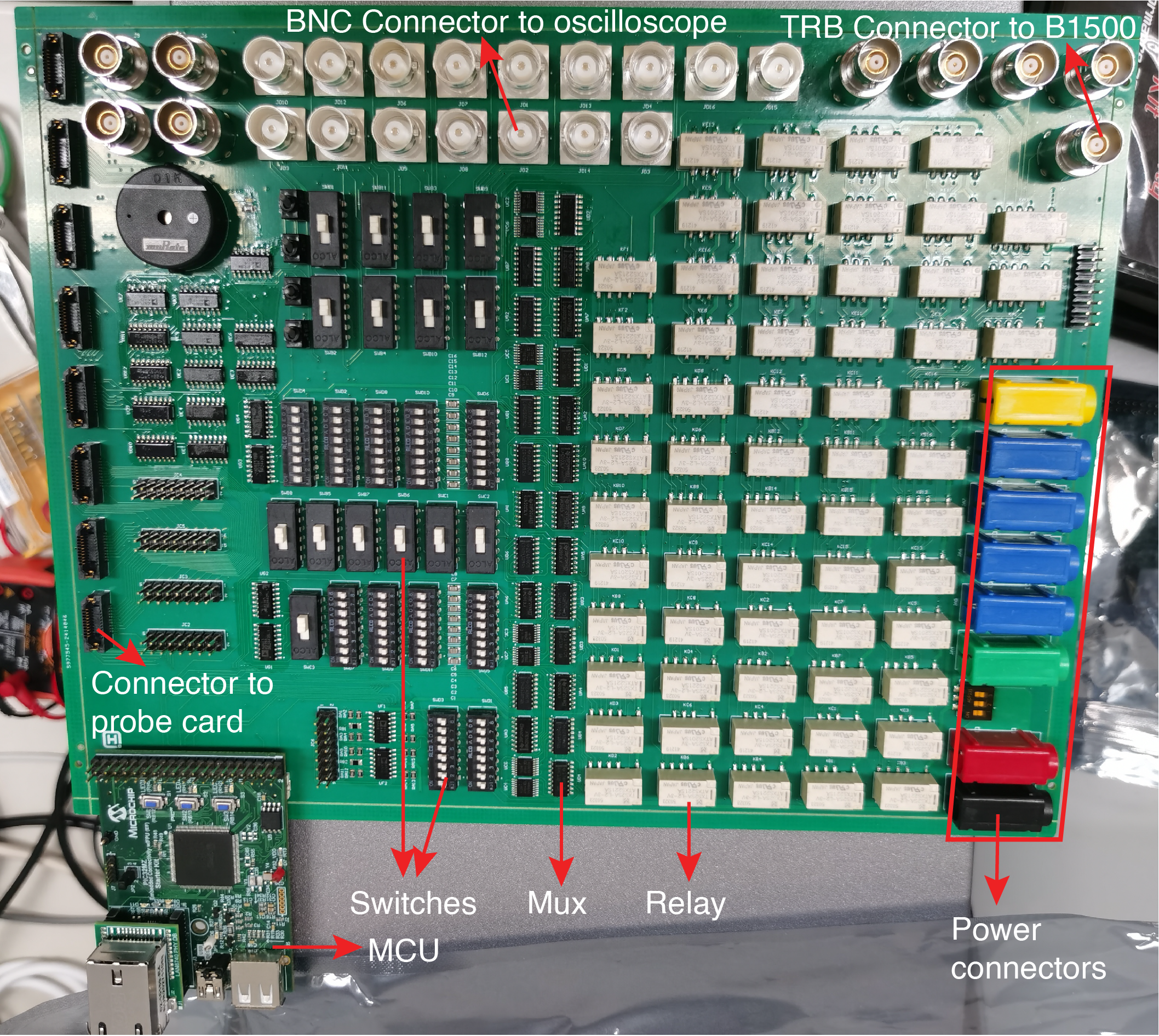}
    \caption{\textbf{PCB-based test board for measurement platform.} The multiplexers (MUXs) and relays on the test board are controlled by a Microchip PIC32 MCU connected with PC. The signals are collected by probe card and transferred through the board to B1500 for analysis. There are 16 connectors for oscilloscope to monitor the charge/discharge process of different analog CAM rows. 
    \can{Please rotate the pic by 180 degrees, and add the lables. Do you have the circuit diagram?}
    }\label{test board}
\end{figure}



\clearpage

\begin{figure}[ht!]%
    \centering
    \includegraphics[width=10.5cm]{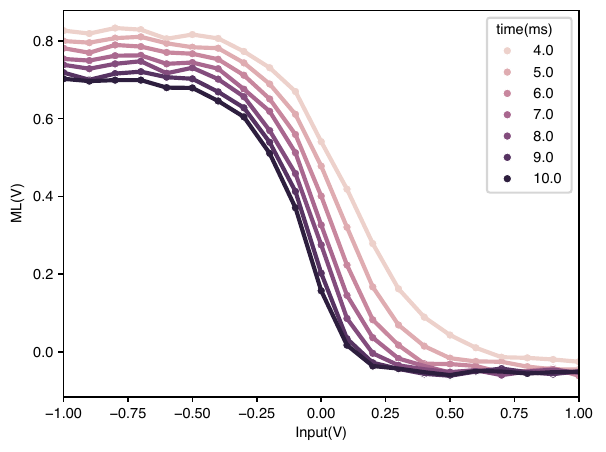}
    \caption{\textbf{Sigmoid-like curves of ML voltage as output.} The ML voltages are sampled at different time points from 4.0 ms to 10.0 ms. The curves get softer (flatter) with less time latency.  }
    \label{curves_different_sampling_time}
\end{figure}

\clearpage

\begin{figure}[ht!]%
    \centering
    \includegraphics[width=6.5cm]{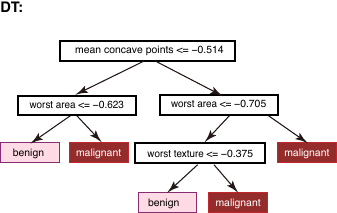}
    \caption{\textbf{Baseline hard decision tree (DT) with sharp binary splits. Features are normalized to [-1, 1].}}
    \label{Baseline DT}
\end{figure}

\clearpage

\begin{figure}[ht!]%
    \centering
    \includegraphics[width=6.5cm]{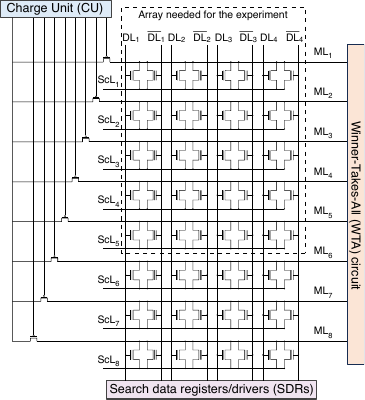}
    \caption{\textbf{Circuit schematic of the analog CAM for the experiment.} The analog CAM array with a winner-take-all (WTA) circuit to select the output that has the highest ML voltage. }
    \label{Circuit schematic}
\end{figure}

\clearpage

\begin{figure}[ht!]%
    \centering
    \includegraphics[width=5.5cm]{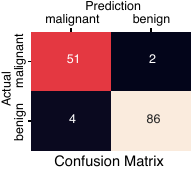}
    \caption{\textbf{Confusion matrix of inference experiment with analog CAM arrays for Breast Cancer dataset.} }
    \label{ConfusionMatrix}
\end{figure}

\clearpage

\begin{figure}[ht!]%
    \centering
    \includegraphics[width=17cm]{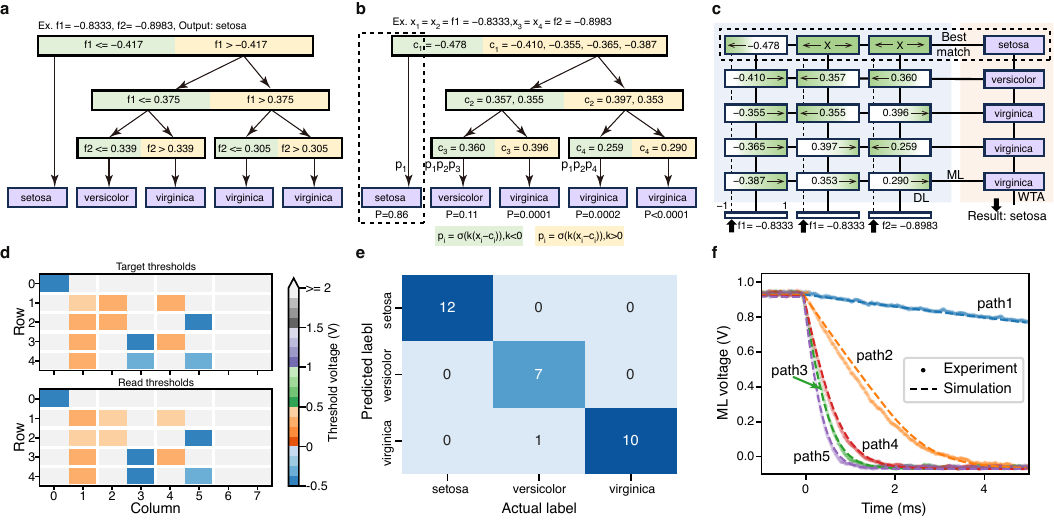}
    \caption{\textbf{Soft decision tree inference experiment with analog CAM arrays for Iris dataset.} 
    \textbf{a}, An illustrative hard tree model for the purpose showing each node makes a binary decision based on a sharp decision boundary. The features are regularized to [-1,1]. 
  \textbf{b}, The demo SDT with the same tree structure of the hard one in (a). 
  Nodes shared by different paths may have slightly different thresholds $c_i$ for the sigmoid gate function; the final output is the leaf with the highest product of node probabilities along its path.   
  \textbf{c}, The SDT implemented in analog CAM with a winner-take-all (WTA) circuit selecting the output that has the highest ML voltage. The``X'' denotes ``always match''. 
  \textbf{d}, The mapping target (upper) and the actual (lower) threshold voltages after programming the analog CAM devices. White indicates a threshold voltage high enough ($\geq \SI{2}{V}$) for the ``always match'' condition. The threshold voltage is defined as the gate voltage when the drain current reaches 10 nA. 
  \textbf{e}, The confusion matrix for the SDT model's prediction performance, where only one sample was misclassified, yielding an accuracy of 97\%.  
  \textbf{f}, An example of ML voltage discharging transient analysis using an oscilloscope for the input in (b). The correct result, corresponding to the highest probability, is obtained from $ML_1$ which represents the leaf node for path1. The dashed lines from the circuit simulation are validated by the experimental data points.  }
    \label{iris_experiment}
\end{figure}

\clearpage

\begin{figure}[ht!]%
    \centering
    \includegraphics[width=9cm]{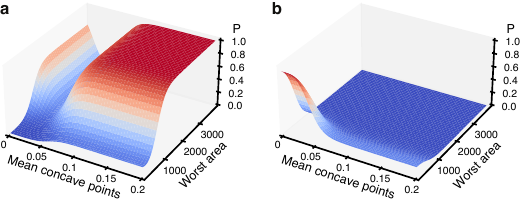}
    \caption{\textbf{The soft decision boundaries of SDT for the Breast Cancer dataset with two features.} The soft decision boundaries in SDT are complex curved surfaces with probabilities, which are for the malignant \textbf{a} and the benign \textbf{b}. The feature not shown is fixed to be an averaged value.}
    \label{Soft decision boundaries}
\end{figure}

\clearpage

\begin{figure}[ht!]%
    \centering
    \includegraphics[width=9cm]{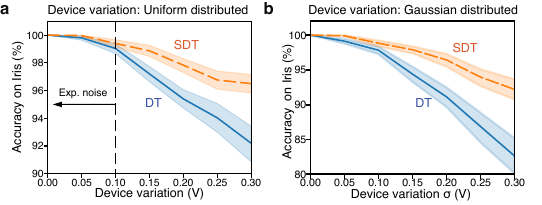}
    \caption{\textbf{Simulation accuracy for Iris under Uniform/Gaussian distributed device variation over 100 times of repetitions.} 
     \textbf{a}, The simulation results of the model's accuracy with 95\% confidence interval on Iris under increasing device threshold variation (uniform distributed). SDT shows superior robustness against threshold variation compared with DT.
     \textbf{b}, The simulation results of the model's accuracy with 95\% confidence interval on Iris under increasing device threshold variation (Gaussian distributed). SDT outperforms DT with increasing standard deviation $\sigma$. The trend is similar to results under uniform distributed device variation. When the $\sigma$ is 0.3V, the accuracy of SDT is 92.2\% $\pm$ 7.7\% compared with 82.6\% $\pm$ 12.8\% of DT .  
    }\label{iris_gaussian}
\end{figure}

\clearpage
\begin{figure}[ht!]%
    \centering
    \includegraphics[width=13cm]{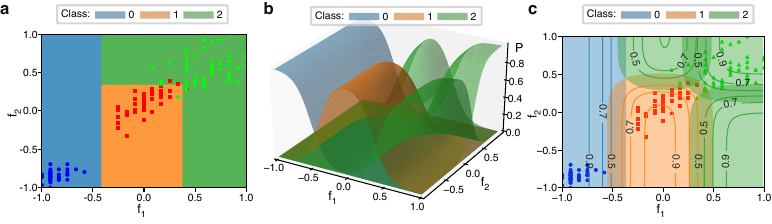}
    \caption{\textbf{Analysis of the DT/SDT decision boundaries on Iris.} The DT used for comparison shares the exact tree structure and feature regularization of SDT, as well as the same row/column architecture for implementation. The main difference is the mapping thresholds of soft models are different through the training considering the soft boundaries in an SDT way. 
  \textbf{a}, The decision area divided by orthogonal boundaries of DT with samples.
  \textbf{b}, The decision surfaces from soft boundaries of SDT.
  \textbf{c}, The contours of probabilities from the decision surfaces in (c) with samples.}
    \label{iris_decision}
\end{figure}

\clearpage

\begin{table}[ht!]
    \centering
    \begin{tabular}{ccccccc} 
    \hline
         Model accuracy&  \multicolumn{3}{c}{DT}&  \multicolumn{3}{c}{SDT}\\ 
    \hline
         Tree depth&  Ideal&  Root attack&  Threshold variation&  Ideal&  Root attack& Threshold variation
\\ 
    \hline
         6&  74.15&  54.79±0.49&  35.23±8.85&  75.66&  67.08±0.18& 74.8±0.88
\\ 
         8&  81.81&  66.43±0.25&  42.44±7.38&  84.31&  78.08±0.23& 83.22±0.4
\\ 
         10&  86.62&  72.94±0.32&  41.99±6.77&  88.12&  83.93±0.24& 87.53±0.18
\\ 
         12&  87.74&  73.51±0.35&  46.36±6.03&  90.1&  87.43±0.18& 89.52±0.19
\\ 
         14&  88.17&  73.93±0.19&  42.47±3.1&  90.77&  88.23±0.25& 90.04±0.21
\\ 
         16&  88.11&  73.94±0.23&  45.27±6.56&  90.79&  88.4±0.13& 90.27±0.23
\\ 
         18&  88.02&  74.05±0.22&  40.37±4.38&  91.15&  88.98±0.13& 90.75±0.17
\\ 
         20&  88.26&  73.92±0.23&  42.99±4.56&  91.26&  89.62±0.12& 90.69±0.18
\\ 
    \hline
    \end{tabular}
    \caption{\textbf{Model's accuracy with increasing tree depth under the root node adversarial attack / threshold variation.} Every result 
    is the average of 10 repetitions considering the randomness.} 
    \label{tab:mnist_accuracy_tree_depth}
\end{table}

\clearpage

\begin{table}[ht!]
\centering
\label{tab:tabular dataset}
\begin{tabular}{cccccc}
\hline
Model accuracy& electricity& bank-marketing& default-of-credit-card-clients& MiniBooNE& MNIST
\\ 
\hline
\#samples& 38474& 10578& 13272& 72998& 60000
\\
\#features& 7& 7& 20& 50& 28*28
\\
 RF (\%) (baseline)& 76.73 [\citenum{Grinsztajn2022tree_outperform}]& 77.96 [\citenum{Grinsztajn2022tree_outperform}]& \textbf{71.58} [\citenum{Grinsztajn2022tree_outperform}]& 89.05 [\citenum{Grinsztajn2022tree_outperform}]&91.88 [\citenum{Random_Forest}]
\\
DT (\%) (this work)& 76.59& 78.22& 68.99& 88.97& 88.26
\\ 
 SDT (\%) (this work)& \textbf{79.08}& 80.06& 70.15& \textbf{91.37}&91.26
\\
 RF (\%) (this work)& 78.55& \textbf{80.62}& 71.47& 90.23&\textbf{96.47}
\\
 SRF (\%) (this work)& 78.35& 80.15& 71.32& 90.36&96.05
\\
\hline
\end{tabular}
\caption{\textbf{Model accuracy for tabular and MNIST datasets.} The number of trees in RF for baseline and our work are both 250 for the four tabular datasets. For MNIST, the baseline RF contains 100 trees while ours own 50 trees. The max tree depth is set to be 4 in [\citenum{Grinsztajn2022tree_outperform}] and we set 6 for all models for tabular dataset. And for [\citenum{Random_Forest}], there is no limit for the maximum tree depth and we limit it to be less than 20 for the MNIST.}
\end{table}

\clearpage

\begin{table}[ht!]
    \centering
    \begin{tabular}{cccc} 
    \hline
         Process&  Latency (ns)&  Energy (nJ)& Reference\\ 
         \hline
         \SI{0.35}{\micro\meter}&  48&  3.62& [\citenum{fish2005high}]\\ 
         \SI{0.18}{\micro\meter}&  80&  1.08& [\citenum{dlugosz2015novel}]\\ 
         \SI{45}{\nano\meter}&  3&  0.07& [\citenum{liu2022cosime}]\\ 
         \hline
    \end{tabular}
    \caption{\textbf{Latency and energy consumption of WTA for 3k inputs.}  }
    \label{tab:WTA}
\end{table}

\clearpage

\begin{table}[ht!]
\centering

\begin{tabular}{l l l l l l}
\hline
\textbf{Model} & \textbf{Accelerator} & \textbf{Process} & \textbf{Power}

\textbf{(mW)} & \textbf{Latency (ns}\textbf{/dec}\textbf{)} & \textbf{Energy (nJ}\textbf{/dec}\textbf{)} \\
\hline
DT & AMD EPYC 7413  & 7 nm & 180 × 10\textsuperscript{3} & 0.6 × 10\textsuperscript{3} & 0.11 × 10\textsuperscript{6} \\
\textbf{SDT} & AMD EPYC 7413 & 7 nm & 180 × 10\textsuperscript{3} & 0.3 × 10\textsuperscript{6} & 54 × 10\textsuperscript{6} \\
\textbf{SDT} & NVIDIA RTX3090 & 8 nm & 350 × 10\textsuperscript{3} & 17.4 × 10\textsuperscript{3} & 6.1 × 10\textsuperscript{6} \\
DT-based & ASIC IMC [\citenum{kang201819}] & 65 nm & 7.1 & 2.7 × 10\textsuperscript{3} & 19.4 \\
DT & TCAM [\citenum{rakka2023dt2cam}] & 16 nm & 247 & 3 & 0.74 \\
DT-based & Memristive 

analog CAM [\citenum{pedretti2021tree}] & 65 nm & 427 & 3 & 1.28 \\
DT & This work & 200 nm & 675 & 10 & 8.78 \\
\textbf{SDT} & This work 

w/o WTA & 200 nm & 675 & 10 & 8.78 \\
\textbf{SDT} & This work

w/ WTA & 200 nm

\& 45 nm & 681 & 13 & 8.85 \\
\hline
\end{tabular}
\caption{\textbf{Latency and energy benchmark.} For the analog CAM implementation, we consider a conservative latency estimation of 10 ns per array based on circuit simulation. The number can be further improved by adapting novel designs based on emerging FeFET technology\cite{Yin2020FeCAM}. 
We also consider the additional peripheral circuitry which is the WTA circuit \cite{WTA}. That causes the latency for SDT slightly higher than DT. }
\label{tab:benchmark}
\end{table}

\clearpage

\bibliography{sn-bibliography}